\documentclass{article} % For LaTeX2e
\usepackage{iclr2024_conference,times}

\usepackage{amsmath,amsfonts,bm}

\def\eqref#1{equation~\ref{#1}}
\def\1{\bm{1}}

\DeclareMathAlphabet{\mathsfit}{\encodingdefault}{\sfdefault}{m}{sl}
\SetMathAlphabet{\mathsfit}{bold}{\encodingdefault}{\sfdefault}{bx}{n}

\usepackage{hyperref}
\usepackage{url}
\usepackage{authblk}
\usepackage{amssymb}
\usepackage{enumitem}
\usepackage{graphicx}
\usepackage{booktabs}
\usepackage{multirow}
\usepackage{subcaption}

\newcommand*{\affaddr}[1]{#1} % No op here. Customize it for different styles.

\title{BOLAA: \underline{B}enchmarking and \underline{O}rchestrating \underline{L}LM-augmented \underline{A}utonomous \underline{A}gents}
\author{{Zhiwei~Liu}$^{\dagger}$\thanks{zhiweiliu@salesforce.com}, 
\text{Weiran~Yao}$^{\dagger}$, 
\text{Jianguo~Zhang}$^{\dagger}$, 
\text{Le~Xue}$^{\dagger}$, 
\text{Shelby Heinecke}$^{\dagger}$,
\text{Rithesh Murthy}$^{\dagger}$,
\text{Yihao Feng}$^{\dagger}$,
\text{Zeyuan Chen}$^{\dagger}$,
\text{Juan Carlos Niebles}$^{\dagger}$,
\text{Devansh Arpit}$^{\dagger}$,
\text{Ran Xu}$^{\dagger}$, 
\text{Phil Mui}$^{\diamond}$,
\text{Huan Wang}$^{\dagger\blacklozenge}$,
\text{Caiming Xiong}$^{\dagger\blacklozenge}$,
\text{Silvio Savarese}$^{\dagger\blacklozenge}$ \\
\affaddr{$^{\dagger}$Salesforce Research, USA}\\
\affaddr{$^{\diamond}$CTO Office, Salesforce, USA\\}
\affaddr{$^{\blacklozenge}$Corresponding Authors: \texttt{\{huan.wang, cxiong, ssavarese\}@salesforce.com}}

}

\iclrfinalcopy % Uncomment for camera-ready version, but NOT for submission.
\begin{document}

\maketitle

\begin{abstract}
The massive successes of large language models (LLMs) encourage the emerging exploration of LLM-augmented Autonomous Agents (LAAs).
An LAA is able to generate actions with its core LLM and interact with environments, which facilitates the ability to resolve complex tasks by  conditioning on past interactions such as observations and actions.
Since the investigation of LAA is still very recent, limited explorations are available. 
Therefore, we provide a comprehensive comparison of LAA in terms of both agent architectures and LLM backbones.
Additionally, we propose a new strategy to orchestrate multiple LAAs such that each labor LAA focuses on one type of action, \textit{i.e.} BOLAA, where a controller manages the communication among multiple agents.
We conduct simulations on both decision-making and multi-step reasoning environments, which comprehensively justify the capacity of LAAs.
Our performance results provide quantitative suggestions for designing LAA architectures and the optimal choice of LLMs, as well as the compatibility of both. 
We release our implementation code of LAAs to the public at \url{https://github.com/salesforce/BOLAA}. 
\end{abstract}

\section{Introduction}
% one more paragraph explain the developments of AI agents.
Recent booming successes of large language models (LLMs)~\citep{openai2023gpt4,touvron2023llama} motivate emerging exploration of employing LLM to tackle various complex tasks~\citep{zhang2023dialogstudio}, amongst which 
\textbf{L}LM-augmented \textbf{A}utonomous \textbf{A}gents (LAAs)~\citep{shinn2023reflexion,madaan2023self,huang2022language,kim2023language,paul2023refiner,yao2023react} stand with most spotlights. 
LAA extends the intelligence of LLM to sequential action executions, exhibiting superiority in interacting with environments and resolving complex tasks via collecting observations. 
To name a few, BabyAGI\footnote{\url{https://github.com/yoheinakajima/babyagi}} proposes an AI-powered task management system, which leverages OpenAI LLM\footnote{\url{https://platform.openai.com/docs/api-reference}} to create, prioritize, and execute tasks. 
AutoGPT\footnote{\url{https://github.com/Significant-Gravitas/Auto-GPT}} is another popular open-source LAA framework that enables the API calling capability of LLMs. 
ReAct~\citep{yao2023react} is a recently proposed LAA method to interact with environments then consecutively generate the next action.
Langchain\footnote{\url{https://github.com/langchain-ai/langchain}} is a recently released open-source framework for developing LAA. 

Due to the initial investigation, LAA is rather under-explored. 
Firstly, the optimal agent architecture is undetermined. 
ReAct~\citep{yao2023react} prompts the agents with pre-defined examples such that the LLM learns to generate the next action via in-context learning.
Moreover, ReAct argues that an agent should have intermediate reasoning steps before action executions.
ReWOO~\citep{xu2023rewoo} introduces additional planning steps for LAA. 
Langchain generalizes the ReAct agent with zero-shot tool usage ability. 
Intrinsically, the optimal architecture of agents should be aligned with both tasks and the associated LLM backbone, which is less explored in the existing works.

Secondly, understanding the efficacy of the existing LLMs in LAA is far from comprehensive. The existing preliminary works only compare the performances of a few LLM backbones.
ReAct adopts the PaLM~\citep{chowdhery2022palm} as the backbone LLM. 
ReWOO employs OpenAI text-davinci-003 model for instruction-tuning Alpaca model~\citep{alpaca} for agent planning. 
MIND2Web~\citep{deng2023mind2web} compares Flan-T5 and OpenAI GPT3.5/4 for generalist web agent. 
Nevertheless, few current works comprehensively compare the performance of LAA with regard to various pre-trained LLMs. 
A very recent work~\citep{liu2023agentbench} releases a benchmark for evaluating LLMs as Agents. Nevertheless, they fail to jointly consider the agent architectures along with their LLM backbones.
Selecting the optimal LLMs from both efficacy and efficiency perspectives advances the current exploration of LAA. 

Thirdly, the increasing complexity of tasks may require the orchestration of multiple agents. 
ReWOO recently identifies that decoupling  reasoning from observation improves the efficiency for LAA. 
In this paper, we argue that as the task complexity increases, especially in open-domain environments, 
it is better to coordinate multiple agents to complete one task. 
For example, regarding the web navigation task, we could employ one \textit{click agent} to interact with clickable buttons and request another \textit{search agent} to retrieve additional resources. 
Nonetheless, there are few works discussing how to orchestrate multiple agents and investigating the impacts of orchestration.

To address these research gaps, this paper proposes to comprehensively compare the performances of LAAs. 
We dive deep into the agent architecture of LAAs and the LLM backbones.
Specifically, we construct agent benchmarks from the existing environments to evaluate the performances of various agent architectures built upon various LLM backbones. 
The tasks in our agent benchmarks are associated with different task complexity levels, which enables the agent performance analyses w.r.t. task complexity.
Those agent architectures are designed to extensively verify the existing design choices. 
Regarding the orchestration of multiple LAAs, we propose a novel LAA architecture BOLAA\footnote{For easy memorizing, we intentionally name it the same as paper title.}, which has a controller module on top of multiple collaborated agents, for enabling the selection and communication between multiple labor LAA.

The contributions of this paper are as follows:
\begin{itemize}[leftmargin=*]
    \item We develop 6 different LAA agent architecture. We combine them with various backbone LLMs to justify the designing intuition of LAA from prompting, self-thinking, and planning.
    We also develop BOLAA for orchestrating multi-agent strategy, which enhances the action interaction ability of solo agents.
    \item We conduct extensive experiments on both decision-making web navigation environment and knowledge reasoning task environment. 
    We report the performance in terms of final sparse rewards and intermediate recalls, which provides qualitative indications for the optimal choice of LAAs as well as their compatible LLMs.
    \item BOLAA on the WebShop environment consistently yields the best performance compared with other LAA architectures. 
    Our results demonstrate that the importance of designing specialist agents to collaborate on resolving complex task, which should be as equally important as training a large LLM with high generalization ability.
\end{itemize}

\section{Related Work}
\subsection{Augmented Language Agent Architecture}
% From search to learn, from earlier to later
% Add the applications like, HuggingGPT, BabyAGI as applications of the proposed agent structure
The completion of a complex task typically entails multiple stages. An agent must possess an understanding of these stages and plan accordingly. Chain-of-Thoughts, also known as CoT~\citep{wei2022chain}, is a groundbreaking work that prompts the agent to deconstruct challenging reasoning tasks into smaller, more manageable steps. On the other hand, ReAct~\citep{yao2023react} proposes leveraging this aptitude for reasoning and action within Language and Learning Models (LLMs) to foster interactive engagement with the environment, such as utilizing the Wikipedia search API, by mapping observations to the generation of reasoning and action traces or API calls in natural language. This agent architecture has given rise to various applications, including HuggingGPT~\citep{shen2023hugginggpt}, Generative Agents~\citep{park2023generative}, WebGPT~\citep{nakano2021webgpt}, AutoGPT~\citep{autogpt23}, BabyAGI~\citep{babyagi23}, and Langchain~\citep{langchain23}. 

However, these approaches neglect to incorporate valuable feedback, such as environment rewards, to enhance the agent's behaviors, resulting in performances that rely solely on the quality of the pre-trained Language and Learning Model (LLM). Self-refine~\citep{madaan2023learning} tackles this limitation by employing a single LLM as a generator, refiner, and provider of feedback, enabling iterative refinement of outputs. However, it is not specifically tailored for real-world task-based interaction with the environment. On the other hand, REX~\citep{murthy2023rex} and RAP~\citep{hao2023reasoning} repurpose the LLM to function as both a comprehensive world model and a reasoning agent. They incorporate Monte Carlo Tree Search for strategic exploration within the vast realm of reasoning with environment rewards. This approach facilitates effective navigation and decision-making in intricate domains. \citet{shinn2023reflexion} presents Reflexion, a framework that equips agents with dynamic memory and self-reflection capabilities, enhancing their reasoning skills. Self-reflection plays a pivotal role, allowing autonomous agents to iteratively refine past actions, make improvements, and prevent repetitive errors. Recently, \citet{yao2023retroformer} proposes a framework, namely Retroformer, which leverages policy gradient optimization to align the agent's behaviors with environment-specific rewards by learning a plug-in retrospective language model.
\subsection{Web Agent}
Web navigation is the foundation for humans to collect information and communicate. 
Before the boom of LLM, previous endeavours~\citep{liu2018reinforcement,shi2017world} already explored how to train web agent in a web simulation environment.
Very recently, a series of works have been devoted to developing LAA to tackle complex web navigation tasks. 
Though action space of web navigation is almost infinite due to numerous available elements online, these action can be divided into a few operation types, such as \textit{click}, \textit{type} and \textit{select}. 
MIND2Web~\citep{deng2023mind2web} collects a web browser data to fine-tune LLM to generate executable actions, which functions as a Web LAA. 
WebAgent~\citep{gur2023real} is able to decompose task instruction into sub-tasks, which directly generates executable python program for web navigation. 
WebArena~\citep{zhou2023webarena} supports realistic tasks simulation for designing Web LAA. 
Langchain and ChatGPT both provide convenient web plugin such that the LLM behaves as Web LAA.
We believe that the web navigation is the next fundamental task for LAA to shine its superiority. 
\subsection{Tool Agent}
The evolution of LLM and their interactions with various tools has been a focal point of recent research. The concept of a ``Tool Agent" encapsulates the idea of LLMs leveraging external tools to enhance their capabilities and solve complex tasks. One of the pioneering works in this domain is the introduction of ``Gorilla"~\citep{patil2023gorilla}.
This model is adept at writing API calls and exhibits the ability to adapt test-time document changes.
% An important contribution of this work is the demonstration of how LLMs can integrate with document retrievers, enabling them to use tools more accurately and stay updated with frequently changing documentation.
Another noteworthy work is the ``ToolLLM" framework~\citep{qin2023toolllm}. 
This open-source framework incorporates LLMs to efficiently engage with a myriad of tools, particularly APIs, to execute intricate tasks. The framework encompasses ToolBench, an instruction-tuning dataset tailored for tool utilization
% , and introduces a Depth-First Search-Based Decision Tree (DFSDT) to amplify the planning and reasoning capabilities of LLMs. The derived model, ToolLLaMA, showcases an impressive capability to carry out complex instructions and generalize to previously unseen APIs, rivaling the performance of models like ChatGPT.
More recently, a paradigm shift in teaching LLMs to use new tools has been discussed in \citep{hsieh2023tool}, which champions the use of tool documentation.
The authors present empirical evidence suggesting that tool documentation offers detailed descriptions of tool usage, which is a more effective and scalable approach. Notably, their research indicates that zero-shot prompts, which are exclusively based on tool documentation, can rival the performance of few-shot prompts. 
% This highlights the potential trajectory for LLMs to seamlessly adapt to a broader array of tools in a zero-shot manner.

\section{Agent Architectures}
In this section, we compare various LAA architectures. 
We first present how to design different solo LAA based on the intuition of existing work. We then present the our orchestration designing of multiple LAAs, \textit{i.e.} BOLAA. 

\subsection{Solo Agents}
\begin{figure}[t!]
    \centering
    \includegraphics[width=0.9\textwidth]{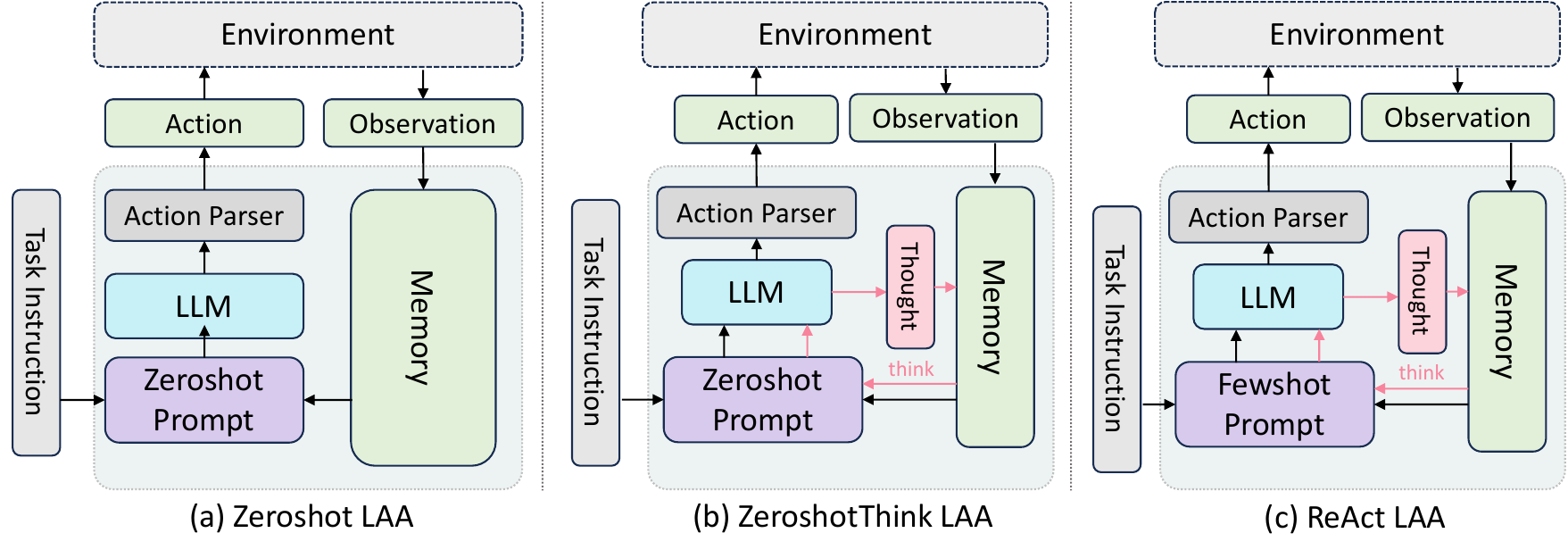}
    \caption{The LAA architectures for Zeroshot-LAA (ZS-LAA), ZeroshotThink LAA (ZST-LAA) and ReAct LAA. ZS-LAA generates actions from LLM with zeroshot prompt. ZST-LAA extends ZS-LAA with self-think. ReAct LAA advances ZST-LAA with fewshot prompt. They all resolve a given task by interacting with environment via actions to collect observations. Better view in colors.}
    \label{fig:zslaa}
\end{figure}

Hereafter, we present 5 different LAAs.
Each type of LAA is able to interact with the environment with its own interaction strategy. 

\textbf{Zeroshot LAA}~(ZS-LAA) 
directly extends the LLM to be action executor.
Specifically, the prompt for LLMs to function as the action executor consists of detailed descriptions for those actions. 
For example, if we prompt LAA to understand the \textit{click} action with ``\textit{click: using this action to click observed [button], the clickable buttons are in [].}", it may behave as a web navigation agent.
We present the architecture of ZS-LAA in Figure~\ref{fig:zslaa}(a). 
The working flow is as follows:
\begin{itemize}[leftmargin=*]
    \item \textit{Initial step}: firstly, the ZS-LAA receives the task instruction and constructs the zeroshot prompt. Then, the LLM layer generates a possible response, which is parsed to output a feasible action.  After that, the observation from environment is appended into the agent memory.
    \item \textit{Working teps}: the agent checks whether the task is finished. If not, ZS-LAA retrieves the previous actions and observations from memory, and constructs the prompts for LLM to generate the next executable actions. ZS-LAA continues the working stage until reaching the maximum steps or completing the task.
\end{itemize}
ZS-LAA is a minimum LAA architecture. It enables the action generation ability of LLM via zeroshot prompt layer, which is easy to generalize to new environments and requires no examples. 

\textbf{ZeroshotThink LAA}~(ZST-LAA) is an extended version of ZS-LAA. Different from ZS-LAA, ZST-LAA  has an additional self-think flow.
The architecture of ZST-LAA is presented in Figure~\ref{fig:zslaa}(b), where we denote the self-think flow as in pink arrow lines. 
Self-think is running in intermediate steps of action generations flow, which enables the Chain-of-Thought (CoT) reasoning ability. 
\begin{itemize}[leftmargin=*]
    \item \textit{Self-think Step}: before generating the next action, ZST-LAA collect observations and previous actions to construct the \textit{think} prompt. Then, the \textit{thought} is stored into memory. 
\end{itemize}
Self-think step is generally useful when given reasoning tasks. 
Note that the think prompt is also in a zero-shot format, such as \textit{``think: using this action to plan your actions and reasoning"}.

\textbf{ReAct LAA} additionally advances ZST-LAA in the prompt layer, where fewshot examples are provided.
The architecture of ReAct LAA is illustrated in Figure~\ref{fig:zslaa}(c). 
ReAct LAA is able to leverage successful running examples to improve the action generation ability of LLM and enhance the environment interaction of LAA, because those fewshot examples endows the in-context learning ability of LLM. 
However, the drawback for ReAct LAA is that, due to the limited context length, fewer token spaces are available after the occupancy of fewshot examples in the prompt. 

\begin{figure}
    \centering
    \includegraphics[width=0.9\textwidth]{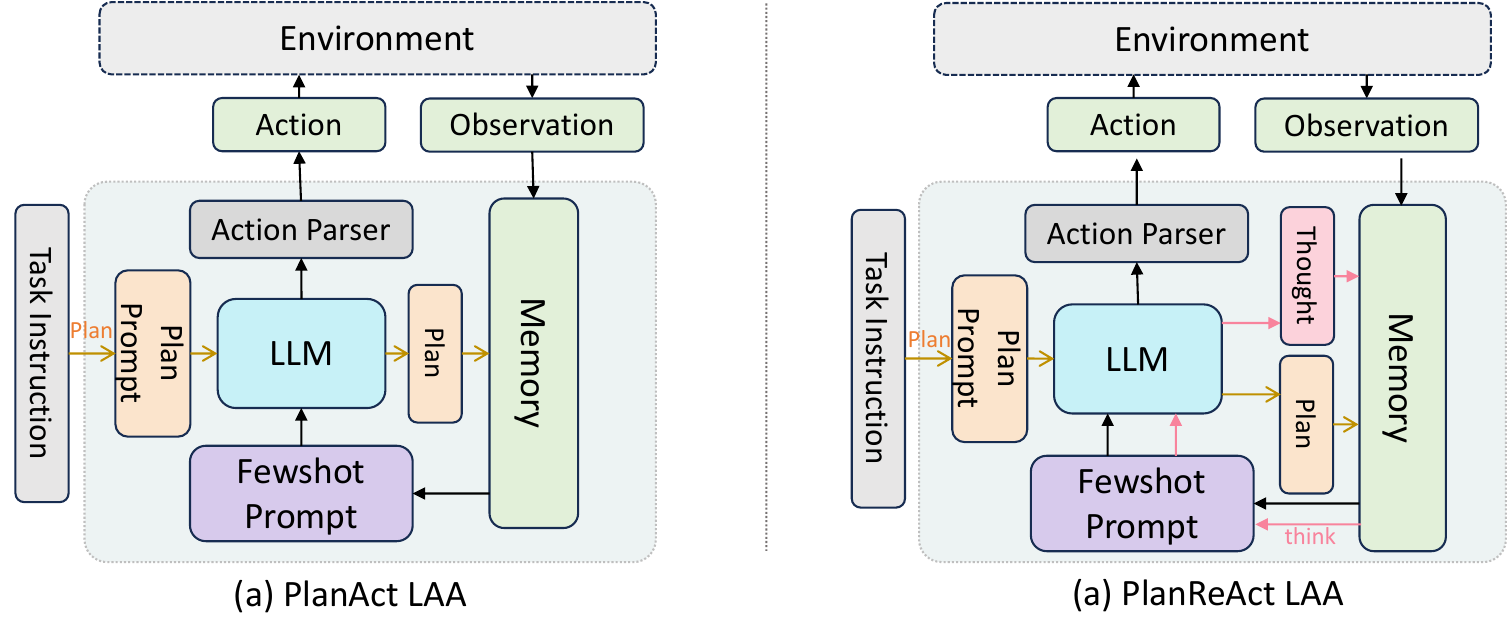}
    \caption{The LAA architectures for PlanAct LAA and PlanReAct LAA. }
    \label{fig:PlanAct}
\end{figure}

\textbf{PlanAct LAA} is designed to facilitate the planning ability of LAA. PlanAct LAA differs from ZS-LAA in two parts: 1) the planning flow and 2) the fewshot prompt. 
The architecture is depicted in Figure~\ref{fig:PlanAct}.
The planning flow is executed before the initial action generation step, which has additional plan prompt to construct the input for the core LLM.
\begin{itemize}[leftmargin=*]
    \item \textit{Planning Step}: PlanAct LAA generates a plan for a given task before interacting with environments. 
    The plan is memorized and will be retrieved to construct prompts. 
\end{itemize}
It is worth noting that the plan prompt in this paper is in fewshot way, which allows LAA to generate plans based on previous successful plans. 

\textbf{PlanReAct LAA} extends PlanAct LAA with additional self-think flow, which also enables the CoT ability. 
The architecture of PlanReAct LAA is presented in Figure~\ref{fig:PlanAct}.
Intuitively, since the Planning flow is executed before the LAA observes the environment, self-think flow alleviates the hallucination incurred from incorrect plans. 

Next, we introduce our multi-agent orchestrating architecture, \textit{i.e.} BOLAA. 

\subsection{BOLAA: Orchestrating Multiple Agents.}
\begin{figure}[!h]
    \centering
    \includegraphics[width=0.9\textwidth]{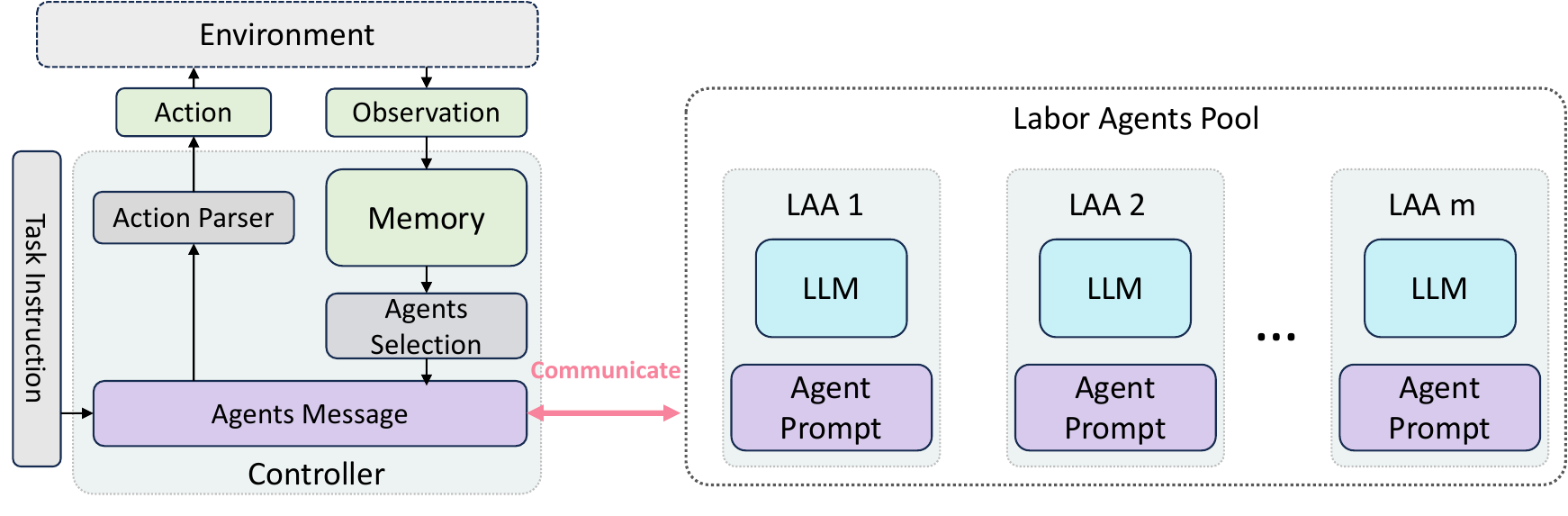}
    \caption{The BOLAA architecture, which employs a controller to orchestrate multiple LAAs.}
    \label{fig:orchestrator}
\end{figure}
Though the success of the existing LLMs in completing various language understanding tasks, plenty of issues are still under-explored, such as the context length constraints, in-context learning and generalization ability, and etc.
Hence, it is challenging to employ a solo LAA to complete all tasks, especially when tasks are of high complexity. 
Therefore, we propose a new agent architecture for orchestrating multiple LAAs, which is illustrated in Figure~\ref{fig:orchestrator}.
BOLAA has two main modules, the labor agents pool and the controller. 
The labor agents pool manages multiple LAAs. 
Each LAA may only focus on generating one type of actions. 
For example, in the web navigation environment, we could establish \textit{click} LAA and \textit{search} LAA. 
In this way, the former only generates the next button to click, while the later only outputs search query, which divides a complex task into feasible tasks. 
The controller is devised to selectively call LAAs from agents pool. 
Controller has the agents selection layer for choosing the most relevant LAA to call. 
Then, the controller constructs the message for the selected LAA and builds the communication. 
After obtaining the response from the labor LAA, the controller parses it to an executable action and then interacts with the environment.
Note that we can also design those labor LAAs to be think/plan agent.
In this way, the self-think and plan work flows are also retained.

\section{Experiment}
\subsection{Environment Benchmark}
We construct the evaluation benchmarks from two environments, \textit{i.e.,} the WebShop~\citep{yao2022webshop} and HotPotQA~\citep{yang2018hotpotqa} with Wikipedia API usage~\citep{yao2023react}.

WebShop is a recently proposed online shopping website environment with 1.18M real-world products and human instructions.
Each instruction is associated with one ground-truth product, and contains attribute requirements, \textit{e.g. I'm looking for a travel monopod camera tripod with quick release and easy to carry, and price lower than 130.00 dollars.}
This instruction includes 3 attribute requirements \textit{i.e.} ``quick release", ``camera tripod" and ``easy carry" attributes.
We define the complexity of an instruction using the number of attribute requirements.
Thus, this instruction example above is of complexity $3$. 
We equally sample 150 instructions regarding each complexity level.
Since we have fewer than 150 instructions for complexity larger than 6, we only include instructions from complexity in $\{1,2,\dots, 6\}$, which sums up to 900 tasks for benchmark evaluation in the WebShop environment.
In the WebShop environment, an agent operates either \textsc{search[query]} or \textsc{click[element]} actions to interact the environment, for evaluating the interactive decision making ability of LAA.
The observation from WebShop is simplified web browser, which includes the clickable buttons and associated page content. 
LAA interacts with the WebShop environment as a web navigation agent.

% \subsubsection{Question Answering (QA)}

% In our daily life, whenever one needs to find an answer to a question out of their knowledge, the default approach is to search and browse through websites. 
HotPotQA with Wikipedia API is another environment considered in this paper, 
which contains multi-hop questions answering tasks that requires reasoning
over two or more Wikipedia passages.
% HotPotQA~\citep{yang2018hotpotqa,yao2023react} environment simulates this task in a virtual world. 
% A wide range of questions are provided to the agent, with the primary objective of browsing through Wikipedia pages to find the most accurate answers to open-ended questions. 
This simulation environment serves as a powerful tool for evaluating the multi-step planning and comprehension capabilities and information retrieval skills of AI models, ensuring they are proficient in sourcing reliable information from vast online resources. With its unique blend of real-world internet browsing scenarios and text analysis, HotpotQA is an invaluable asset for the advancement of augmented large language agent systems.
In HotPotQA environment, an agent has three types of actions, \textit{i.e.}, \textsc{search[entity]}, \textsc{lookup[string]} and \textsc{finish[answer]} to interact with HotPotQA environment.
HotPotQA environment aims at evaluate the knowledge reasoning ability of LAA.
We randomly sample 100 questions from easy, medium and hard levels, which constitutes the final 300 benchmark questions for evaluating LAAs.

% \paragraph{Implementation Detail}

% \paragraph{Documentation}

% \begin{enumerate}[leftmargin=*] \itemsep0em
% \item \textbf{Observation}.
% \item \textbf{Action API}.
% \item \textbf{Reward}.
% \item \textbf{Termination}. 
% \end{enumerate}

\subsection{Evaluation Metrics}
We mainly use the \textit{reward} score in each environment to evaluate the performances of LAAs. 
In the WebShop environment, the reward is defined as the attribute overlapping ratio between the bought item and ground truth item. 
In HotPotQA environment, the reward is defined as the F1 score grading between agent answer and ground-truth answer. Additionally, we develop the \textit{Recall} performance for WebShop environment, which is defined as 1 if the ground truth item is retrieved and 0 if not during one task session. 
The Recall is reported as the average recall scores across all tasks in WebShop environment. 

\subsection{LLM Utilization}
The core component of LAA is the LLM backbone. We compare different LLMs with various choices of model size and context length. 
We reported the results w.r.t. open LLM models such as fastchat-3b, vicuna-3b/13b/33b~\citep{zheng2023judging}, Llama-2-7b/13b/70b\footnote{All Llama-2 models are -chat-hf version.}~\citep{touvron2023llama}, MPT-7b/30b~\citep{MosaicML2023Introducing}, xgen-8k-7b, longchat-16k-7b/13b and OpenAI API LLMs, including text-davinci-003, gpt-3.5-turbo and
gpt-3.5-turbo-16k. 

\subsection{Decision-making Simulation}
\begin{table}[t]
\caption{Average reward in the WebShop environment. Len denotes the maximum context length. \textbf{Bold} results denote the best results in one row, \textit{i.e.} best LAA architecture w.r.t. one LLM. \underline{Underline} results denote the best performance in one column, \textit{i.e.} best LLM regarding one LAA architecture.}
\label{tab:webshop_reward}
\begin{tabular}{l|c|cccccc}
\toprule
\multirow{2}{*}{LLM} & \multirow{2}{*}{Len.} & \multicolumn{6}{c}{LAA Architecture}                                                          \\ \cmidrule(l){3-8} 
                     &                       & ZS     & ZST    & ReAct           & PlanAct               & PlanReAct       & BOLAA           \\ \hline
{fastchat-t5-3b}        & {2k}                    & 0.3971 & 0.2832 & 0.3098          & 0.3837                & 0.1507          & \textbf{0.5169} \\
{vicuna-7b}            & {2k}                    & 0.0012 & 0.0002 & \textbf{0.1033} & 0.0555                & 0.0674          & 0.0604          \\
{vicuna-13b}           & {2k}                    & 0.0340  & 0.0451 & 0.1509          & 0.3120                & 0.4127          & \textbf{0.5350}  \\
{vicuna-33b}           & {2k}                    & 0.1356 & 0.2049 & 0.1887          & 0.3692                & 0.3125          & \textbf{0.5612} \\
{llama-2-7b}           & {4k}                    & 0.0042 & 0.0068 & 0.1248          & 0.3156                & 0.2761          & \textbf{0.4648} \\
{llama-2-13b}          & {4k}                    & 0.0662 & 0.0420  & 0.2568          & \underline{\textbf{0.4892}} & 0.4091          & 0.3716          \\
{llama-2-70b}          & {4k}                    & 0.0122 & 0.0080  & 0.4426          & 0.2979                & 0.3770           & \textbf{0.5040}  \\
{mpt-7b-instruct}      & {8k}                    & 0.0001      & 0.0001      & 0.0573          & 0.0656                & \textbf{0.1574} & 0.0632          \\
{mpt-30b-instruct}     & {8k}                    & 0.1664 & 0.1255 & 0.3119          & {0.3060}                & 0.3198          & \textbf{0.4381} \\
{xgen-8k-7b-instruct}  & {8k}                    & 0.0001      & 0.0015 & 0.0685          & 0.1574                & 0.1004          & \textbf{0.3697} \\
{longchat-7b-16k}      & {16k}                   & 0.0165 & 0.0171 & 0.069           & 0.0917                & 0.1322          & \textbf{0.1964} \\
{longchat-13b-16k}     & {16k}                   & 0.0007 & 0.0007 & 0.2373          & 0.3978                & \textbf{0.4019} & 0.3205          \\ \hline
{text-davinci-003}      & {4k}  & 0.5292       & 0.5395       & \underline{0.5474} & 0.4751 & 0.4912       & \textbf{0.6341}       \\
{gpt-3.5-turbo}          & {4k}  & 0.5061       & 0.5057       & 0.5383       & 0.4667 & 0.5483       & \underline{ \textbf{0.6567}} \\
{gpt-3.5-turbo-16k} & {16k} & \underline{0.5657} & \underline{0.5642} & 0.4898       & 0.4565 & \underline{0.5607} & \textbf{0.6541}       \\ \bottomrule
\end{tabular}
\end{table}

In this section, we present and compare the decision-making performances of LAAs in the WebShop environment. 
The performance regarding the average reward is reported in Table~\ref{tab:webshop_reward}.
The agent prompts are constructed based on the maximum context length of different LLM models. 
Regarding BOLAA, we devise one search LAA and one click LAA to generate search query and click elements, respectively.  
We have the following observation:
\begin{itemize}[leftmargin=*]
    \item BOLAA performs the best compared with the other LAA architectures, especially when built on the high performing LLMs. BOLAA is able to actively select the appropriate LAA and yield qualitative communication, which stabilizes the action generation.  
    We observe that BOLAA, when paired with a 3b fastchat-t5 LLM, performs comparably to other LAA architectures with more powerful LLMs.
    The superiority of BOLAA indicates that orchestrating multiple smaller-sized LAAs is  a better choice if the computing resources are limited. 
    This further exemplifies the potential for fine-tuning multiple smaller-sized specialised LAAs rather than fine-tuning one large generalized LAA.
    \item Pairing the LLM with the optimal LAA architecture is crucial. For example, Llama-2-13b performs best under PlanAct LAA arch while Llama-2-70b performs best under the BOLAA arch. Also, Longchat-13b-16K performs best when using PlanAct and PlanReAct, which may indicate the extraordinary planning ability of longchat-13b-16k models. 
    \item Increasing the context length alone may not necessarily improve the LAA performances. For example, when comparing longchat-13b-16k with llama-2-13b models, the latter yields better performances though with less context length. By checking the running log of those LAAs, we observe more occurrence of hallucinated generation when the LAA runs for more steps, which in the end degrades the benefits of longer context.
    \item A powerful LLM is able to generalize under the zeroshot LAA arch.  The best performance of OpenAI API-based models are actually under ZS and ZST arch. This indicates the great potential of developing a generic LAA with powerful LLM. 
    Actually, this is currently what open-source projects are working towards, directly calling OpenAI API and tuning the zeroshot agent prompt instead. 
    Our benchmark results quantitatively justify that using only a ZS LAA can already achieve comparable or even better performances than LAA arch with additional Plan or Self-think flow. However, for other less powerful LLMs, fewshot prompts are necessary for LAAs. 
    \item Plan flow generally improves the performances when the agent is built on open-source LLMs. 
    By comparing the performances of ReAct, PlanAct and PlanReAct, we observe a performance gain on most LLM cases when using plan flow. However, planning and thinking require the LLM to be able to reason in steps, which may be challenging for small size LLMs. 
    For example, fastchat-t5-3b performs above average on ZS LAA arch. But the performance degrades by a large margin under PlanReAct arch.
\end{itemize}

\begin{table}[]
    \centering
    \caption{Average recall in the WebShop environment. Len denotes the maximum context length. \textbf{Bold} results denote the best results in one row, \textit{i.e.} best LAA architecture w.r.t. one LLM. \underline{Underline} results denote the best performance in one column, \textit{i.e.} best LLM regarding one LAA architecture.}
    \label{tab:webshop_recall}
    \begin{tabular}{l|c|cccccc}
\toprule
\multirow{2}{*}{LLM} & \multirow{2}{*}{Len.} & \multicolumn{6}{c}{LAA Architecture}                                                          \\ \cmidrule(l){3-8} 
                     &                       & ZS     & ZST    & ReAct           & PlanAct               & PlanReAct       & BOLAA           \\ \hline
fastchat-t5-3b         & 2k          & 0.3533 & 0.3122 & 0.3800 & 0.3700  & 0.3722    & \textbf{0.3867} \\
vicuna-7b              & 2k          & 0.0833 & 0.0500 & 0.3600 & 0.3233  & 0.3278    & \textbf{0.3522} \\
vicuna-13b             & 2k          & 0.0867 & 0.0644 & 0.3622 & 0.3444  & 0.2367    & \textbf{0.3700} \\
vicuna-33b             & 2k          & 0.3600 & 0.3411 & 0.3822 & 0.3733  & 0.3567    & \textbf{0.3956} \\
llama-2-7b             & 4k          & 0.0678 & 0.0311 & 0.3744 & 0.3400  & 0.3578    & \textbf{0.3856} \\
llama-2-13b            & 4k          & 0.2856 & 0.2211 & 0.3844 & 0.3278  & 0.3500    & \underline{\textbf{0.4078}} \\
llama-2-70b            & 4k          & 0.3344 & 0.3244 & 0.3789 & 0.3400  & 0.3600    & \textbf{0.4011} \\
mpt-7b-instruct        & 8k          & 0.0144 & 0.0322 & \textbf{0.3644} & 0.3200  & 0.3400    & 0.3600 \\
mpt-30b-instruct       & 8k          & 0.2973 & 0.3372 & 0.3333 & 0.3575  & 0.3412    & \textbf{0.3900} \\
xgen-8k-7b-instruct    & 8k          & 0.0667 & 0.1400 & 0.3711 & 0.3400  & 0.3278    & \textbf{0.3800} \\
longchat-7b-16k        & 16k         & 0.1344 & 0.1856 & 0.3644 & 0.3622  & 0.3622    & \textbf{0.3811} \\
longchat-13b-16k       & 16k         & 0.0756 & 0.0867 & 0.3678 & 0.3467  & 0.3471    & \textbf{0.3789} \\
\hline
text-davinci-003       & 4k          & 0.3800 & \underline{0.3856} & 0.3767 & 0.3711  & \underline{0.3889}    & \textbf{0.3956} \\
gpt-3.5-turbo          & 4k          & \underline{0.3889} & 0.3756 & \textbf{0.3933} & \underline{0.3789}  & 0.3867    & 0.3929 \\
gpt-3.5-turbo-16k-0613 & 16k         & 0.3856 & 0.3833 & \underline{\textbf{0.4011}} & 0.3756  & 0.3811    & 0.3933 \\
\bottomrule
\end{tabular}
\end{table}

% Besides the overall reward performance, we also select the 
We also report the intermediate Recall performances for all LAAs, which are illustrated in Table~\ref{tab:webshop_recall}.
Recall is mainly related to the search action. 
High recall performances indicate that the LAA is capable of generating a precise search query.
High recalls usually lead to better rewards. But they are not tightly related. 
For example, Llama-2-70b has a recall performance of nearly 0.3344 on ZS LAA, which is comparable to the best LAA.
However, the reward performance in Table~\ref{tab:webshop_reward} of ZS LAA Llama-2-70b is only 0.0122. 
The reason is that generating the search query requires a different LLM ability from generating the correct click action, where the latter is more challenging. 
Another observation is that our proposed BOLAA generally performs the best on all LLMs, which indicates that separating the search agent from the click agent improves the accuracy of the search action, leading to a higher recall value.

\begin{figure}[h!]
    \centering
    \begin{subfigure}[b]{0.49\textwidth}
        \includegraphics[width=\textwidth]{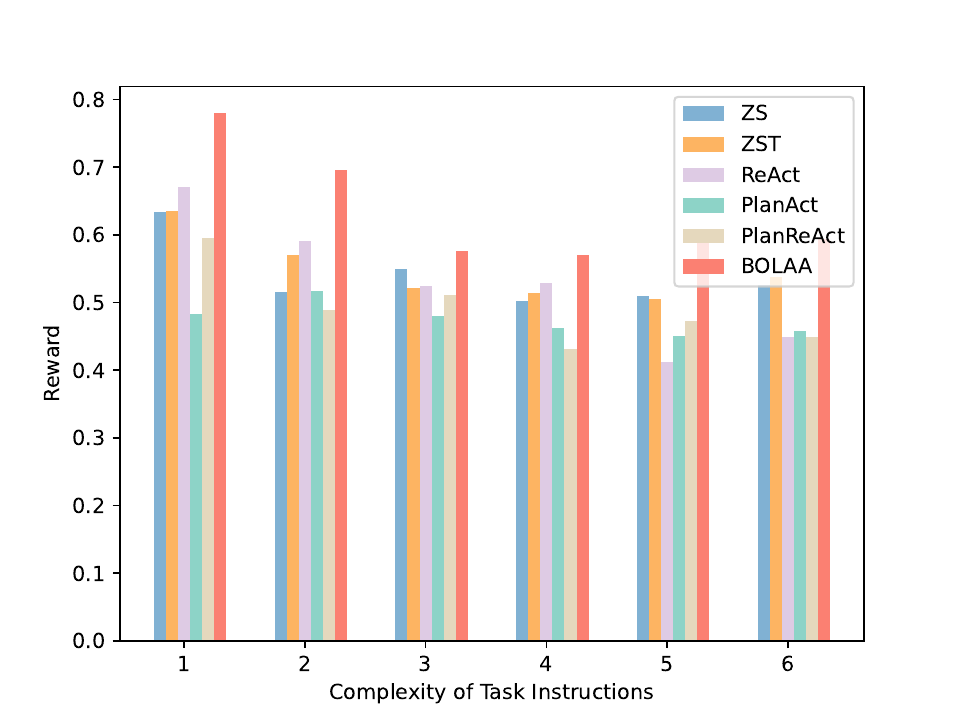}
    \caption{text-davinci-003}
    \label{fig:text-davinci-003-reward}
    \end{subfigure}
    \begin{subfigure}[b]{0.49\textwidth}
        \includegraphics[width=\textwidth]{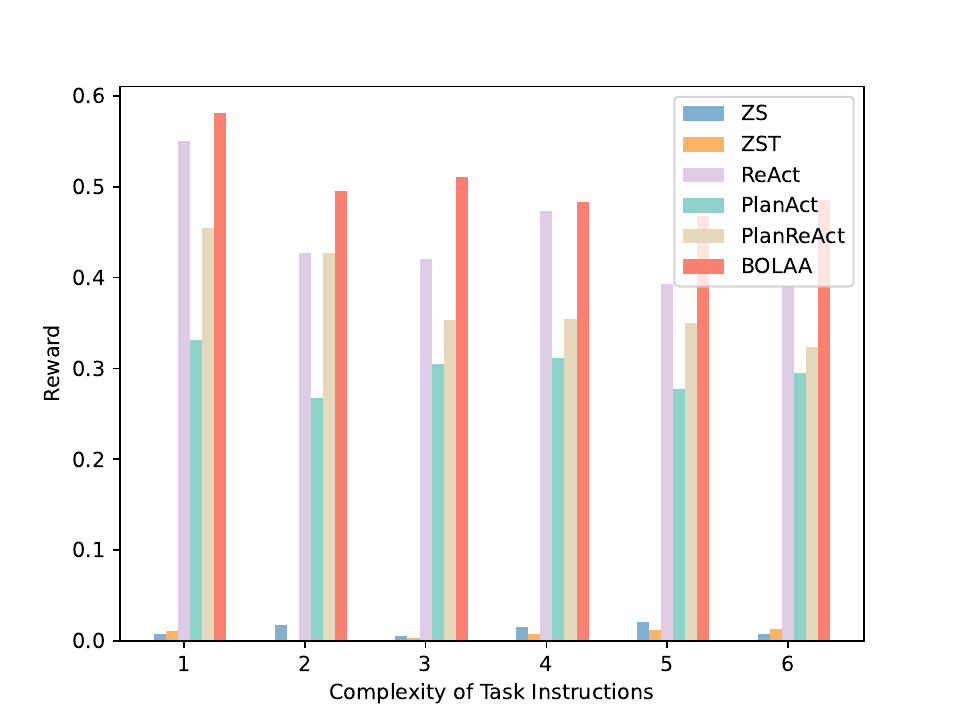}
    \caption{Llama-2-70b}
    \end{subfigure}
    \caption{The reward w.r.t. task complexity in WebShop. Each bar represents one LAA.}
    \label{fig:complexity_reward}
\end{figure}

\textbf{LAA performance w.r.t. Complexity}. 
After the overall performances of those LAAs and LLMs are compared, we conduct more details investigation of the performance w.r.t. the task complexity. 
Due to the space limitation, we only report the performance of text-davinci-003 and llama-2-70b. The reward performance is illustrated in Figure~\ref{fig:complexity_reward}.
The BOLAA model consistently performs better on all complexity levels. 
We also observe the degraded performances when the task complexity is increased, which follows the intuition. 
\begin{figure}[h!]
    \centering
    \begin{subfigure}[b]{0.49\textwidth}
        \includegraphics[width=\textwidth]{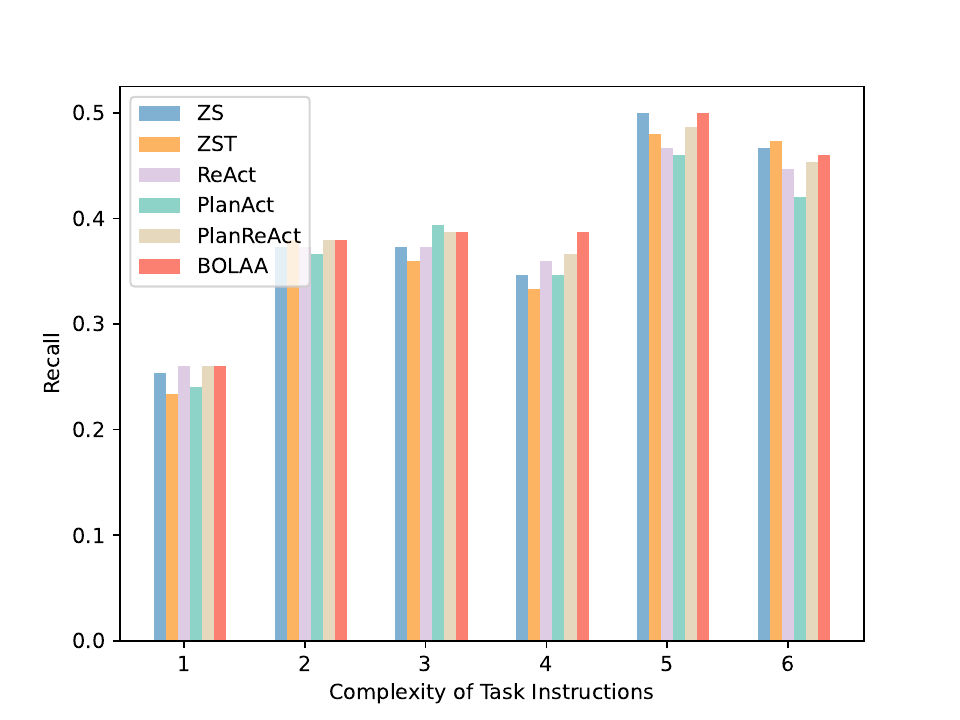}
    \caption{text-davinci-003}
    \label{fig:text-davinci-003}
    \end{subfigure}
    \begin{subfigure}[b]{0.49\textwidth}
        \includegraphics[width=\textwidth]{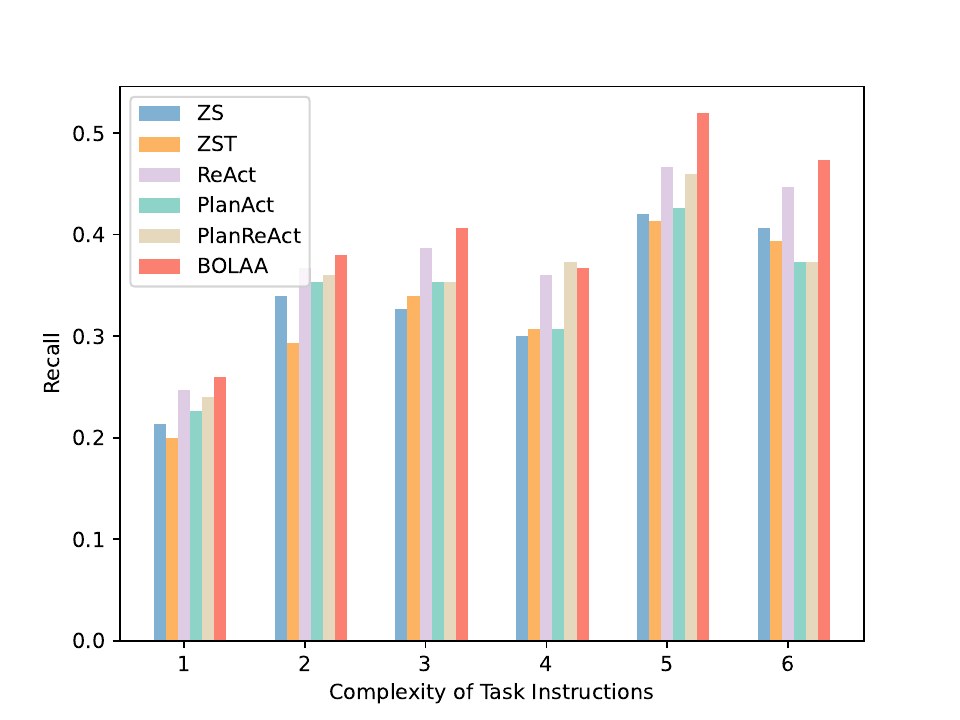}
    \caption{Llama-2-70b}
    \end{subfigure}
    \caption{The recall w.r.t. task complexity in WebShop. Each bar represents one LAA.}
    \label{fig:complexity_recall}
\end{figure}
Surprisingly, we find out that further increasing the complexity of tasks greater than 4 will not further degrade the performances. The reason is that the recall performance increases when the task is of higher complexity, which we demonstrated in Figure~\ref{fig:complexity_recall}. This is due to the fact that high-complexity task instruction provides more additional context information for the LAA. As such, the \textit{search} action can be more specific and accurate under high complexity levels. 

\subsection{Knowledge Reasoning Simulation}
We benchmark on the HotPotQA environment to evaluate the multi-step reasoning ability of LAAs. 
Since the available search, lookup and finish operations are all related to knowledge reasoning in this environment and hard to separate, we therefore leave the BOLAA arch for future work and only compare the performance on other agent arch.
The results are in Table~\ref{tab:hotpot_reward}. 
In general, ReAct agent arch achieves the best performances, which can be interpreted in twofold.
Firstly, fewshot prompt is necessary to enable the action generation and reasoning ability for LAA, especially when experimenting with those small-size language models. 
Secondly, comparing ReAct, PlanAct, and PlanReAct, we would conclude that planning flow of LAA hinders performance the in knowledge reasoning environment and tasks. 
The reason is that knowledge reasoning tasks require contextualized information to conduct reasoning, whereas planning flow is executed ahead of interactions.
Thus, those generated plans tend to lead to more hallucination of LAA.
Thirdly, regarding this knowledge reasoning task, model size is much more important than the context length. 
Large-sized model has better abilities in reasoning, thus performing better. 
Additionally, the superior reasoning ability of OpenAI gpt-3.5 models is again verified. 
We also observe the best performance of Llama-2-70b on all open-source LLMs, which suggests that potential future fine-tuning can be applied on Llama-2 models. 

\begin{table}[!h]
    \centering
    \caption{Average reward in the HotPotQA environment. Len denotes the maximum context length. \textbf{Bold} results denote the best results in one row, \textit{i.e.} best LAA architecture w.r.t. one LLM. \underline{Underline} results denote the best performance in one column, \textit{i.e.} best LLM regarding one LAA architecture.}
    \label{tab:hotpot_reward}
    \begin{tabular}{l|c|ccccc}
\toprule
\multirow{2}{*}{LLM} & \multirow{2}{*}{Len.} & \multicolumn{5}{c}{LAA Architecture}                                                          \\ \cmidrule(l){3-7} 
                     &                       & ZS     & ZST    & ReAct           & PlanAct               & PlanReAct            \\ \hline
fastchat-t5-3b         & 2k          & 0.0252          & 0.0067 & 0.0692          & \textbf{0.1155} & 0.0834 \\
vicuna-7b              & 2k          & \textbf{0.1339} & 0.0797 & 0.0318          & 0.0868          & 0.0956 \\
vicuna-13b             & 2k          & 0.1541          & 0.0910 & \textbf{0.2637} & 0.1754          & 0.2075 \\
vicuna-33b             & 2k          & 0.2180          & 0.2223 & \textbf{0.2602} & 0.1333          & 0.2016 \\
llama-2-7b             & 4k          & 0.0395          & 0.0207 & \textbf{0.2624} & 0.1780          & 0.1417 \\
llama-2-13b            & 4k          & 0.1731          & 0.2313 & \textbf{0.2521} & 0.2192          & 0.2177 \\
llama-2-70b      & 4k & 0.2809                 & 0.3207                  & \textbf{0.3558}           & 0.1424
        & 0.1797          \\
mpt-7b-instruct        & 8k          & 0.0982          & 0.0483 & \textbf{0.1707} & 0.1147          & 0.1195 \\
mpt-30b-instruct       & 8k          & 0.1562          & 0.2141 & \textbf{0.3261} & 0.2224          & 0.2315 \\
xgen-8k-7b-instruct    & 8k          & 0.1502          & 0.1244 & \textbf{0.1937} & 0.1116          & 0.1096 \\
longchat-7b-16k        & 16k         & 0.0791          & 0.0672 & \textbf{0.2161} & 0.1296          & 0.0971 \\
longchat-13b-16k       & 16k         & 0.1083          & 0.0562 & \textbf{0.2387} & 0.1623          & 0.1349 \\
\hline
text-davinci-003 & 4k & \underline{0.3430}           & \underline{0.3304}            & \underline{\textbf{0.4503}}     & \underline{0.3577}                & \underline{0.4101}                  \\
gpt-3.5-turbo          & 4k          & \textbf{0.3340} & 0.3254 & 0.3226          & 0.2762          & 0.3192 \\
gpt-3.5-turbo-16k-0613 & 16k         & \textbf{0.3027} & 0.2264 & 0.1859          & 0.2113          & 0.2251 \\
\bottomrule
\end{tabular}
\end{table}

\begin{figure}[h!]
    \centering
    \begin{subfigure}[b]{0.49\textwidth}
        \includegraphics[width=\textwidth]{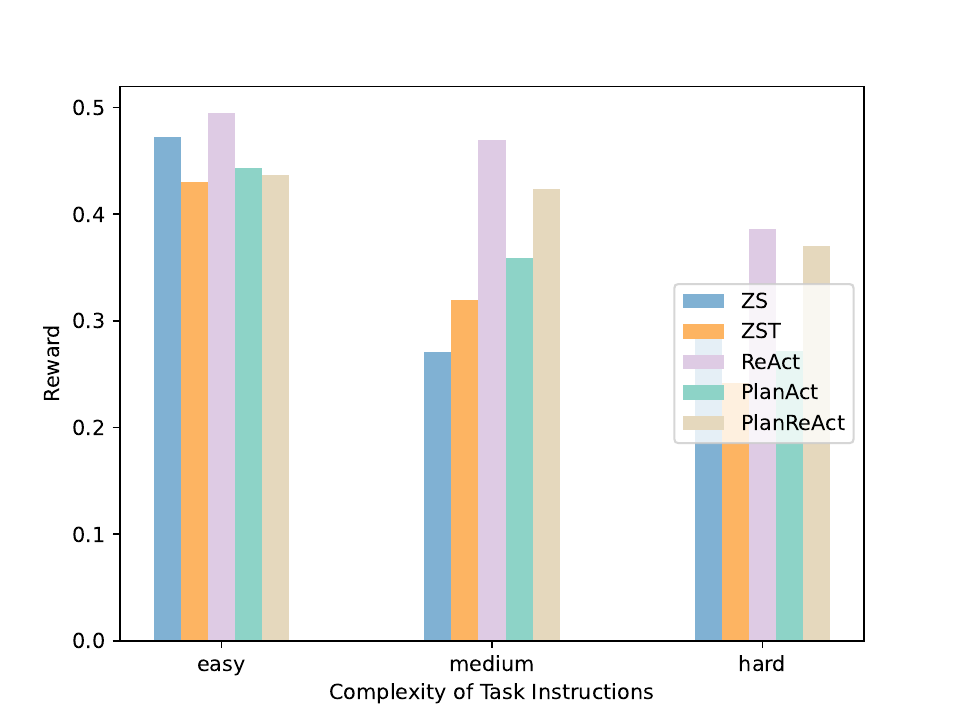}
    \caption{text-davinci-003}
    \label{fig:gpt-3.5-turbo}
    \end{subfigure}
    \begin{subfigure}[b]{0.49\textwidth}
        \includegraphics[width=\textwidth]{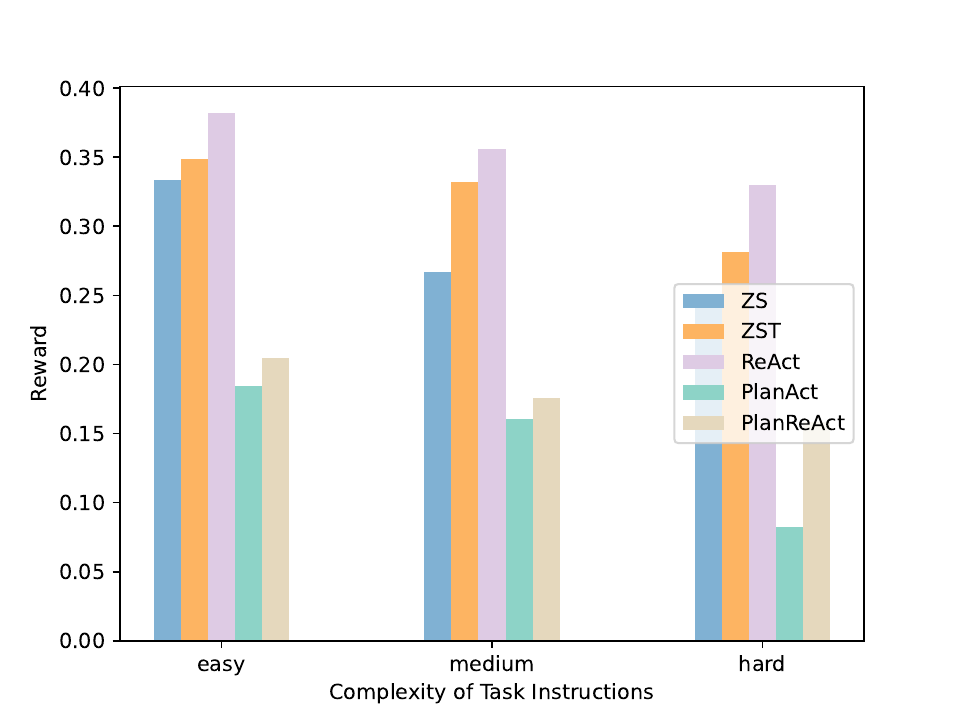}
    \caption{Llama-2-70b}
    \label{fig:llama-2-70b}
    \end{subfigure}
    \caption{The reward w.r.t. complexity level in HotPotQA. Each bar represents one LAA.}
    \label{fig:hotpot_complexity_reward}
\end{figure}

\textbf{LAA performance w.r.t. Complexity}. 
Since we have easy, medium, and high level tasks, we compare the performance of Llama-2-70b and  regarding different levels of complexity, as illustrated in Figure~\ref{fig:hotpot_complexity_reward}. 
We observe degrading performance if increasing the complexity of tasks. 
In HotPotQA tasks, the hardness is defined as the question answer hops.
Therefore, hard question requires more context understanding and reasoning ability of LAA.
Though OpenAI text-davinci-003 model consistently outperforms Llama-2-70b on all levels of complexity, their difference is of smaller margin in hard questions.
% Since easy questions are all from training data and we have no idea whether OpenAI fine-tuned their model on easy
Since hard questions requires more resoning efforts, we can conclude that Llama-2-70b  posses comparable reasoning ability with text-davinci-003. 
\section{Conclusion and Future Work}
In this paper, we systematically investigate the performances of various LAA architecture paired with different LLM backbones. 
We also provide one novel orchestrating method for multiple agents, \textit{i.e.} BOLAA. 
The benchmarking results provide experimental justification for the LAA investigation and verify the potential benefits of BOLAA architecture. 
During the investigation, we also identify the challenge of designing BOLAA architecture for environments with compounding actions. 
In the future, we will explore whether we can harness LLMs in the controller such that selection and communication with labor agents is also fully autonomous.
We will continue developing more LAA architectures and include more LLMs and environments for evaluations. 

% \clearpage
\newpage
\bibliography{reference}
\bibliographystyle{iclr2024_conference}

\end{document}